%% file: main.tex
\documentclass{article} %
\usepackage{iclr2027_conference,times}

\input{math_commands.tex}

\usepackage{hyperref}
\hypersetup{pdftitle={DeltaSplat: Iterative Gaussian Refinement for Pose-Free Feed-Forward 3D Gaussian Splatting}, pdfauthor={Chanung Park, Seunghyeon Song, Joo Chan Lee, Eunbyung Park, Jong Hwan Ko}}
\usepackage{url}

\usepackage{graphicx}
\usepackage{booktabs}
\usepackage{makecell}
\usepackage{amsmath,amssymb}
\usepackage{caption} %
\usepackage{subcaption} %
\usepackage{xcolor}
\ifdefined\pdfsuppressptexinfo\pdfsuppressptexinfo=-1\fi %
\newcommand{\tablefont}{\normalsize\renewcommand{\arraystretch}{1.15}} %
\newsavebox{\qualimg} %
\newlength{\qcell}\newlength{\qgap}\newlength{\qcolh}
\newsavebox{\rtkimg}\newlength{\rtkcell}\newlength{\rtkgap}
\newcommand{\rtk}[2]{\includegraphics[width=\rtkcell]{Figures/re10k/re10k_s#1_#2.jpg}}
\newcommand{\rtkrow}[1]{\hbox{\rtk{1}{#1}\hspace{\rtkgap}\rtk{2}{#1}\hspace{\rtkgap}\rtk{3}{#1}\hspace{\rtkgap}\rtk{4}{#1}}}
\newcommand{\rtklabel}[1]{\hbox{\parbox[b][\rtkcell][c]{20pt}{\centering\rotatebox{90}{\small\begin{tabular}{@{}c@{}}#1\end{tabular}}}}}
\newsavebox{\tabA}\newsavebox{\tabB}\newlength{\tabAw}\newlength{\tabBw}\newlength{\tabdh} %
\makeatletter\newcommand{\setstretchpt}[1]{\edef\arraystretch{\strip@pt\dimexpr#1\relax}}\makeatother
\newcommand{\qgridcol}[2]{\includegraphics[height=\qcolh,trim=#1 7 #2 7,clip]{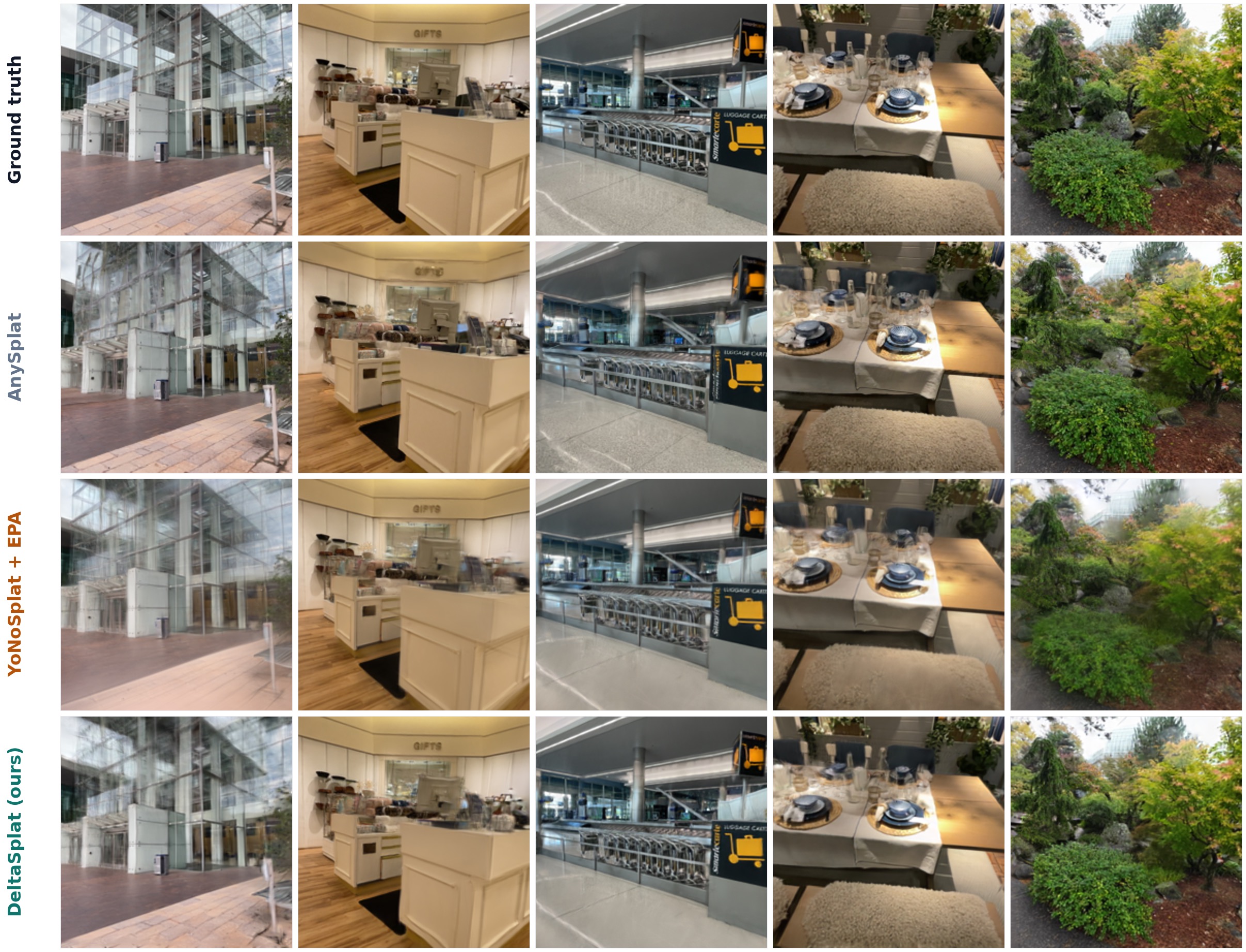}}
\newcommand{\qscenezero}{\vbox{\hsize\qcell\offinterlineskip
  \hbox{\includegraphics[width=\qcell]{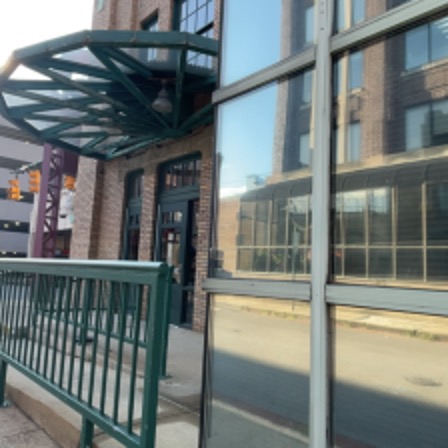}}\vskip\qgap
  \hbox{\includegraphics[width=\qcell]{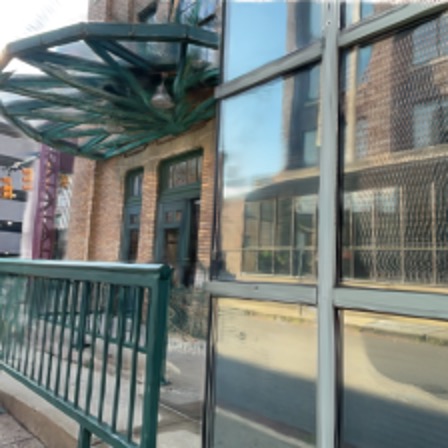}}\vskip\qgap
  \hbox{\includegraphics[width=\qcell]{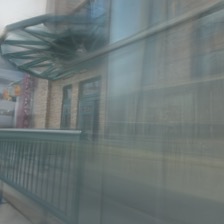}}\vskip\qgap
  \hbox{\includegraphics[width=\qcell]{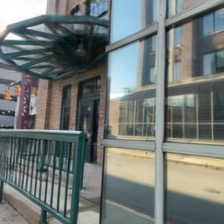}}}}
\newcommand{\qrowlabel}[1]{\parbox[c][0.25\ht\qualimg][c]{20pt}{\centering\rotatebox{90}{\small\begin{tabular}{@{}c@{}}#1\end{tabular}}}}

\makeatletter
\g@addto@macro\@maketitle{%
  \vskip -0.1in%
  \centering%
  \captionsetup{type=figure}%
  \includegraphics[width=\textwidth]{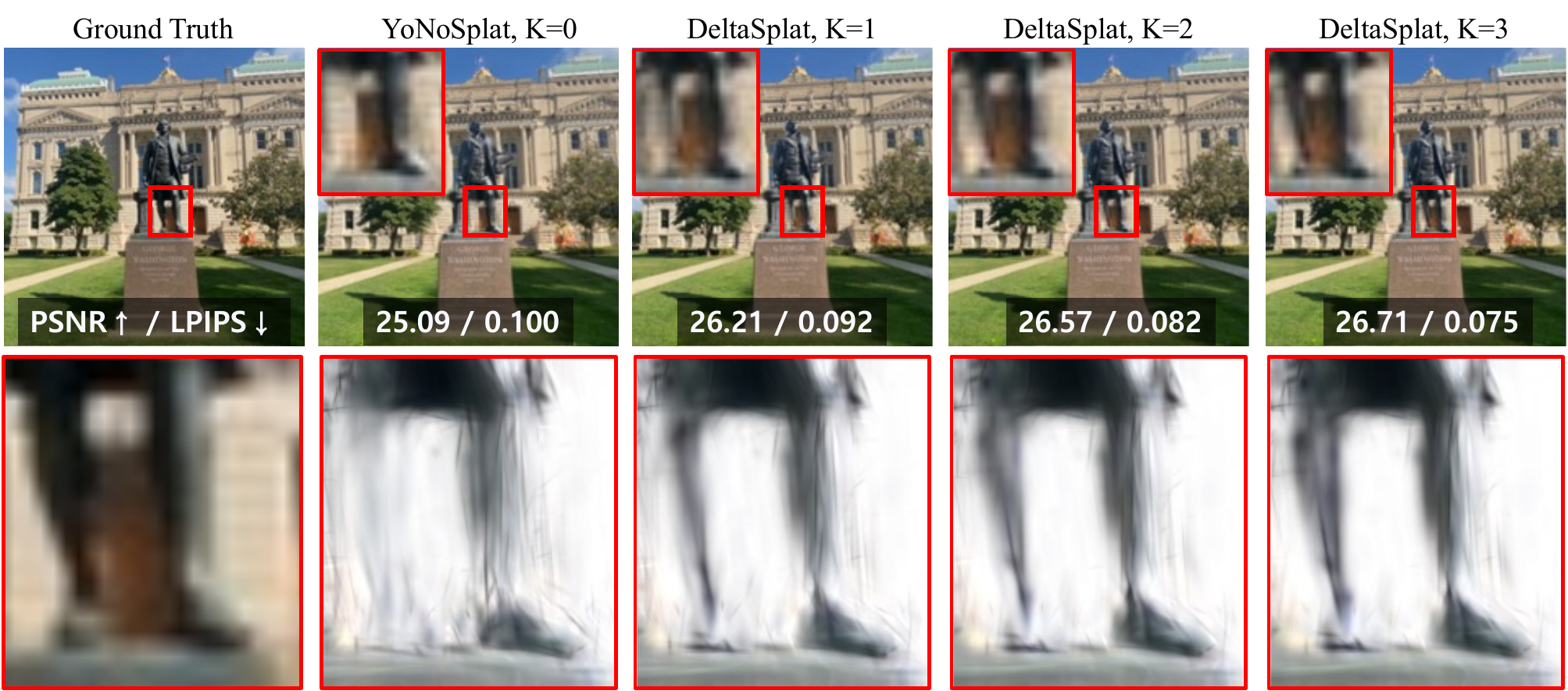}%

  \caption{Iterative Gaussian refinement on a DL3DV scene
in the pose-free setting. $K$ denotes the number of refinement
iterations applied. Starting from the feed-forward output of the
YoNoSplat backbone ($K{=}0$), each DeltaSplat iteration
progressively recovers structures missing from the initial
prediction: the statue's legs, absent at $K{=}0$, gradually
emerge in both the rendered views (top, red box) and the
underlying Gaussian geometry (bottom, with the ground-truth
image crop as reference), with monotonic gains in
PSNR and LPIPS.}%
  \label{fig:teaser}%
  \vskip 0.05in%
  }%
\makeatother

\title{DeltaSplat: Iterative Gaussian Refinement \\
for Pose-Free Feed-Forward \\ {3D} {G}aussian Splatting}

\author{Chanung Park$^{1}$, Seunghyeon Song$^{1}$, Joo Chan Lee$^{2}$, Eunbyung Park$^{3,*}$, Jong Hwan Ko$^{1,*}$ \\
$^{1}$Department of Electrical and Computer Engineering, Sungkyunkwan University \\
$^{2}$Department of Artificial Intelligence, Sungkyunkwan University \\
$^{3}$Department of Artificial Intelligence, Yonsei University
}

\iclrfinalcopy %
\begin{document}

\maketitle
\begingroup\renewcommand\thefootnote{*}\footnotetext{Corresponding authors.}\endgroup
\lhead{}\renewcommand{\headrulewidth}{0pt} %
\captionsetup{skip=6pt} %

\begin{abstract}
Pose-free feed-forward 3D Gaussian Splatting (3DGS) reconstructs a
scene from sparse, unposed images in a single network pass,
removing the need for camera calibration and per-scene
optimization. However, camera estimation errors propagate into the
predicted Gaussians and compound the geometric and photometric
inaccuracies of single-pass prediction. To correct these errors,
we introduce DeltaSplat, a lightweight Gaussian refinement module
for pose-free feed-forward 3DGS. It iteratively renders the current
Gaussians at the input context views and predicts per-Gaussian
updates from the resulting residuals. A 2D residual alone,
however, underdetermines the 3D correction. DeltaSplat therefore
conditions each update on per-pixel Pl\"ucker rays and rendered
depth as a soft geometric prior. A dual-branch convolutional
mixer efficiently encodes these inputs, and per-attribute heads
decode the fused features into position, opacity, and color
updates. The
module adds only ${\sim}2.2\%$ parameters to the backbone and
remains fully feed-forward at inference. On DL3DV, DeltaSplat reaches 26.64\,dB PSNR in the pose-free
setting, improving its state-of-the-art backbone by 1.75\,dB and
surpassing even baselines supplied with ground-truth cameras;
consistent gains hold across 6--24 views and all camera regimes.
\end{abstract}
\newpage %

\section{Introduction}\label{sec:intro}

3D Gaussian Splatting (3DGS)~\citep{kerbl20233dgs} has rapidly emerged as a powerful representation for novel-view synthesis, combining an explicit set of anisotropic Gaussian primitives with differentiable rasterization to achieve high-quality, real-time rendering. Despite its rendering efficiency, conventional 3DGS requires accurate camera calibration and thousands of per-scene optimization steps to reconstruct every new scene. Feed-forward 3DGS~\citep{charatan2024pixelsplat,chen2024mvsplat,xu2025depthsplat,
ye2025noposplat,ye2026yonosplat} addresses this reconstruction bottleneck by amortizing scene-specific optimization into a learned model that directly predicts 3D Gaussians from a sparse set of input images in a single network pass. This paradigm substantially reduces reconstruction time while leveraging learned geometric and appearance priors to handle sparse observations.

Despite removing per-scene optimization, most early feed-forward 3DGS models~\citep{chen2024mvsplat, xu2025resplat, kang2026multi} still assume that accurate camera parameters are available. 
However, the dependence on accurate camera poses substantially limits the practical applicability of feed-forward reconstruction. Although calibrated rigs and controlled capture systems can provide reliable camera parameters, casually captured image collections generally contain neither known extrinsics nor precise intrinsics. Camera poses can be recovered using external sensors or SfM and SLAM pipelines, but these procedures introduce additional preprocessing cost and can be unreliable for sparse, weakly overlapping, or unstructured observations. Moreover, even small pose errors perturb the multi-view correspondences and projection geometry used to construct the Gaussians, leading to noticeable degradation in reconstruction quality. When pose estimation becomes the dominant computational cost or failure point, it also undermines the primary efficiency advantage of feed-forward 3DGS.

To address the aforementioned issue, recent pose-free feed-forward approaches jointly recover a scene representation and its camera geometry from unposed images. AnySplat~\citep{jiang2025anysplat} employs a geometry transformer
to predict 3D Gaussian primitives together with the camera intrinsics and extrinsics of all input views in a single forward pass. NoPoSplat~\citep{ye2025noposplat} anchors one input camera coordinate system as a canonical frame and directly predicts the Gaussians of all views in this shared space, avoiding explicit transformations based on externally estimated poses. YoNoSplat~\citep{ye2026yonosplat} predicts local Gaussians and camera parameters for each view and aggregates them into a global representation using either predicted or provided poses, while its mixed-pose training strategy enables a single model to support posed, unposed, calibrated, and uncalibrated inputs.

Pose-free reconstruction, however, does not eliminate camera uncertainty; it internalizes camera estimation within the reconstruction model. Errors in the predicted cameras perturb the rays and coordinate
transformations used to place the Gaussians, producing the
systematic geometric displacement, ghosting, and blurred
appearance seen in the unrefined prediction of
Figure~\ref{fig:teaser}. These errors interact with the geometric and photometric inaccuracies of single-pass Gaussian prediction, which provides no explicit feedback mechanism for assessing and revising its output. Refining the predicted Gaussian representation is therefore necessary, but deriving a geometrically reliable correction from unposed observations remains challenging.

Rendering residuals provide a potentially useful correction signal, as they directly expose discrepancies between the current reconstruction and the observed context images. ReSplat~\citep{xu2025resplat}, for example, exploits pixel- and feature-space rendering errors to recurrently refine Gaussians under calibrated cameras. In the pose-free setting, however, uncertainty in the predicted camera geometry corrupts the correspondence between image-space residuals and individual Gaussians, thereby destabilizing the inferred 3D updates. Reliable Gaussian refinement under such camera uncertainty therefore requires additional geometric cues to disambiguate the residual-to-Gaussian mapping.

To address this challenge, we introduce DeltaSplat, a lightweight residual refinement module for pose-free feed-forward 3DGS. DeltaSplat iteratively compares the current renderings with the input context images and predicts Gaussian updates from the resulting 2D residuals. To lift these image-space errors into reliable 3D corrections, it jointly incorporates the Pl\"ucker rays~\citep{10.1098/rstl.1865.0017} and rendered depth associated with each Gaussian, providing a soft geometric prior for refining its position and appearance. 
Furthermore, a dual-branch convolutional mixer encodes those fine-grained full-resolution features while processing most features at reduced resolution, enabling more precise refinement with lower computational complexity. 
The encoded features are processed by parameter-specific heads to predict residual updates, which are then mapped back to the corresponding Gaussians to refine their position, opacity, and color. Repeating this process for three shared-weight iterations progressively improves the Gaussian representation. DeltaSplat adds only ${\sim}2.2\%$ to the backbone parameters and requires no per-scene optimization at inference time.

Experiments on DL3DV~\citep{ling2024dl3dv} demonstrate that DeltaSplat substantially outperforms YoNoSplat~\citep{ye2026yonosplat}, a state-of-the-art pose-free feed-forward 3DGS method. With six context views, predicted poses, and known intrinsics, DeltaSplat achieves 26.64\,dB PSNR, improving YoNoSplat by \textbf{+1.75}\,dB under the same evaluation protocol, and surpassing every baseline evaluated with ground-truth cameras. The margins persist at 12 and 24 views and in the fully uncalibrated setting, and DeltaSplat likewise improves the backbone by about $2$\,dB when ground-truth cameras are supplied. These results demonstrate the effectiveness of geometrically guided residual refinement for high-quality pose-free reconstruction without per-scene optimization.

\section{Related Work}\label{sec:related}

\begin{figure}[t]
\centering
\includegraphics[width=\linewidth]{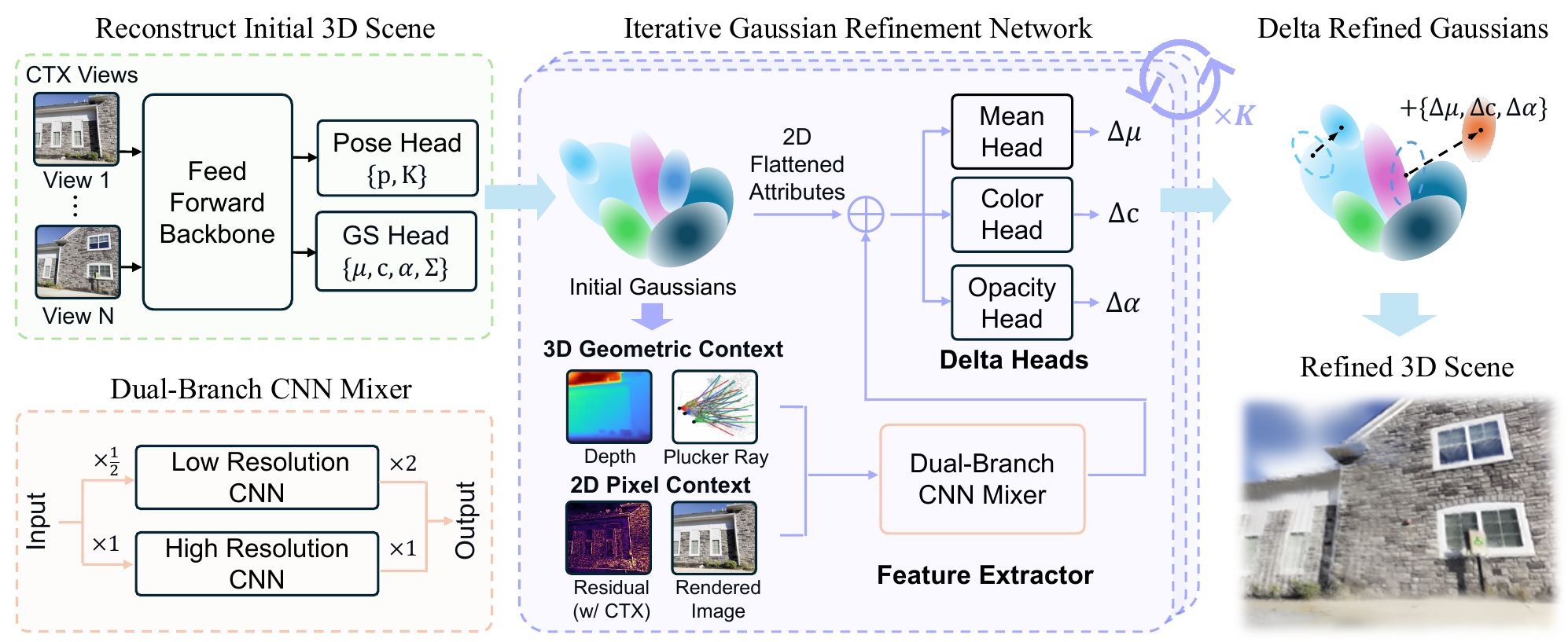}
\caption{\textbf{Method overview.} The backbone predicts cameras
and initial Gaussians; at each iteration, the refinement module
encodes the render residual with the Pl\"ucker rays and the
rendered depth, and decodes the per-Gaussian update
$\{\Delta\mu, \Delta c, \Delta\alpha\}$, repeated $K{=}3$ times
with shared weights.}
\label{fig:method}
\end{figure}

\subsection{Feed-Forward 3DGS}\label{sec:bg-ff}
3D Gaussian Splatting (3DGS)~\citep{kerbl20233dgs} models 3D scenes as explicit, anisotropic Gaussians optimized on hundreds of calibrated images, whose camera poses are
estimated via Structure-from-Motion~\citep{7780814} (SfM). Although it yields high-fidelity renderings, obtaining them requires accurate camera poses and thousands of per-scene optimization steps. Subsequent efforts improve both training and rendering efficiency~\citep{10.1145/3680528.3687694, HansonSpeedy, 10.1007/978-3-031-72980-5_10, lee2025optimized, chen2025dashgaussian, Ren_2026_CVPR, dong2026turbogs}, yet optimizing each scene from scratch remains inherently time-consuming. To overcome this bottleneck, feed-forward (FF) 3DGS amortizes per-scene optimization into a single forward pass, predicting Gaussians directly from sparse context images (typically 2--12 views). Recent architectures inject specialized inductive biases into this feed-forward prediction: pixelSplat~\citep{charatan2024pixelsplat} utilizes epipolar attention, MVSplat~\citep{chen2024mvsplat} and MVSGaussian~\citep{liu2025mvsgaussian} incorporates cost volumes, and DepthSplat~\citep{xu2025depthsplat} leverages pre-trained depth priors. While these approaches achieve sub-second inference, they address only the optimization overhead and remain strictly dependent on ground-truth camera poses.

\subsection{Pose-Free Feed-Forward 3DGS}\label{sec:posefree}
To eliminate the reliance on ground-truth cameras, pose-free FF~3DGS 
leverage geometric foundation models~\citep{wang2024dust3r,leroy2024mast3r,wang2025vggt,yang2026pi3} to construct 3D representations directly from uncalibrated images. Specifically, early methods map Gaussians onto explicit 3D points~\citep{smart2024splatt3r} or relative canonical frames~\citep{ye2025noposplat, huang2025spfsplat}. Although these methods eliminate extrinsic pose supervision, they still require known camera intrinsics. Recent frameworks achieve high reconstruction fidelity even in fully uncalibrated settings. For instance, AnySplat~\citep{jiang2025anysplat} voxelizes pixel-wise primitives for dense captures, YoNoSplat~\citep{ye2026yonosplat} aggregates per-view local Gaussians to improve scalability with intrinsic conditioning module, and 2Xplat~\citep{jeong2026twoxplat} decouples camera estimation and Gaussian synthesis into specialized experts. However, predicting full camera geometry together with Gaussian attributes inevitably compounds pose inaccuracies with feed-forward synthesis limits, resulting in suboptimal 3D representations.

\subsection{Refining Feed-Forward Gaussians}\label{sec:bg-refine}
Standard feed-forward models operate open-loop, leaving a quality gap relative to per-scene optimization. InstantSplat~\citep{fan2024instantsplat} reduces this gap with test-time optimization, but its iterative fitting incurs substantial latency. Diff3R~\citep{liu2025diff3r} instead trains the network to predict an initialization that downstream test-time optimization can refine efficiently, yet it too relies on this iterative optimization at inference. Learned refiners avoid this overhead by directly predicting parameter updates: SplatFormer~\citep{chen2025splatformer} and ReSplat~\citep{xu2025resplat} refine Gaussians using rendering residuals, yet both strictly assume ground-truth poses to establish exact 2D to 3D correspondences.

In pose-free settings, camera inaccuracies distort rays and decouple 2D residuals from true 3D locations. While concurrent work GIFSplat~\citep{chen2026gifsplat} addresses this spatial ambiguity using generative diffusion priors, we aim to resolve it geometrically by leveraging per-pixel ray directions and rendered depth alongside color residuals to guide Gaussian updates, effectively compensating for pose errors and enhancing visual fidelity.

\section{Method}\label{sec:method}

\subsection{Preliminaries}\label{sec:prelim}

Given a set of $N$ images
$\{I_i \in \mathbb{R}^{H \times W \times 3}\}_{i=1}^{N}$, we
partition them into context views
$\mathcal{C} = \{1,\dots,N_\text{ctx}\}$ and target
views $\mathcal{T} = \{N_\text{ctx}+1,\dots,N\}$. A pose-free
feed-forward predictor $f_\theta$
directly maps the context images to a global 3D scene
representation and its associated camera geometry:
\begin{align}
 \big(\mathcal{G}_\text{ff},\; \{p_i\}_{i \in \mathcal{C}}\big)
&=f_\theta\big(\{I_i\}_{i \in \mathcal{C}}\big),\\
\mathcal{G}_\text{ff}
&= \big\{(\mu_g,\, \alpha_g,\, q_g,\, s_g,\,
c_g^{(0)})\big\}_{g=1}^{G}.
\end{align}
The predicted collection $\mathcal{G}_\text{ff}$ comprises $G$
3D Gaussians, where each primitive $g$ is characterized by its
spatial mean $\mu_g \in \mathbb{R}^3$, opacity
$\alpha_g \in (0,1)$, and anisotropic shape defined via rotation
$q_g \in \mathbb{S}^3$ and scale
$s_g \in \mathbb{R}^3_{+}$. Additionally,
each Gaussian carries a degree-0 spherical harmonic (SH)
coefficient $c_g^{(0)}$ to capture its view-independent color.
These primitives are
pixel-aligned with each context view, resulting in a total count
of $G = N_\text{ctx} H W$. Each camera $p_i = [K_i, R_i, t_i]$ encapsulates the intrinsics
$K_i \in \mathbb{R}^{3 \times 3}$ and the extrinsic pose
composed of a rotation $R_i \in \mathrm{SO}(3)$ and a
translation $t_i \in \mathbb{R}^3$; both the intrinsics and the
pose are predicted by $f_\theta$, though either may instead be
supplied when calibration is available.

To synthesize a novel view, a differentiable rasterizer
$\mathcal{R}$ renders the Gaussian collection $\mathcal{G}$ at a
target camera $p_i$ expressed in the coordinate frame of the
prediction:
\begin{equation}
\hat{I}_i = \mathcal{R}\big(\mathcal{G};\, p_i\big)
\qquad \text{for } i \in \mathcal{T}.
\end{equation}

\subsection{Iterative Gaussian Refinement}\label{sec:overview}

To bridge the quality gap caused by single-pass prediction and camera estimation errors, we introduce a refinement module $h_\phi$ that
progressively improves the Gaussian representation, as
illustrated in Figure~\ref{fig:method}. We define
$\mathcal{G}^{(k)}$ as the Gaussian set at the $k$-th refinement
step, initializing the process with the backbone's feed-forward
output, $\mathcal{G}^{(0)} = \mathcal{G}_\text{ff}$.

At each iteration $k \in \{0,\dots,K{-}1\}$, the current state
$\mathcal{G}^{(k)}$ is rendered from the predicted context
cameras $p_i$, yielding a rendering residual $r_i^{(k)}$ that
captures the pixel-wise discrepancy between the current
rendered image $\hat{I}_i^{(k)}$ and the input context image
$I_i$:
\begin{align}
\hat{I}_i^{(k)} &= \mathcal{R}\big(\mathcal{G}^{(k)};\,p_i\big),\\
r_i^{(k)} &= I_i - \hat{I}_i^{(k)}.
\end{align}

Conditioned on the residuals
$r_i^{(k)} \in \mathbb{R}^{H \times W \times 3}$, the
renderings $\hat{I}_i^{(k)} \in \mathbb{R}^{H \times W \times 3}$,
the rendered depths $d_i^{(k)} \in \mathbb{R}^{H \times W}$, and
the Pl\"ucker ray maps~\citep{10.1098/rstl.1865.0017}
$\ell_i \in \mathbb{R}^{H \times W \times 6}$ of the context
cameras, together with the current
Gaussian state, the module predicts one update per Gaussian:
\begin{equation}
\Delta^{(k)} = h_\phi\big(\big\{r_i^{(k)},\,\hat{I}_i^{(k)},\,
d_i^{(k)},\,\ell_i\big\}_{i \in \mathcal{C}},\;
\mathcal{G}^{(k)}\big),
\end{equation}
where each per-Gaussian update $\Delta_g^{(k)} \in \mathbb{R}^7$
consists of a 3D position shift $\Delta\mu_g \in \mathbb{R}^3$,
an opacity adjustment $\Delta\alpha_g \in \mathbb{R}$, and a
color correction $\Delta c_g^{(0)} \in \mathbb{R}^3$. The Gaussian
state is then updated as
\begin{equation}
\mathcal{G}^{(k+1)} = \mathcal{G}^{(k)} \oplus \Delta^{(k)},
\end{equation}
where $\oplus$ adds each predicted increment to its respective
attribute, while the scales $s_g$ and rotations $q_g$ remain
fixed throughout the refinement, preserving the local structure
established by the backbone. As in prior iterative estimation methods~\citep{teed2020raft}, the module weights are shared across all $K$ iterations, and repeating this process $K$ times yields the refined set
$\mathcal{G}^{(K)}$.

\subsection{Geometry-Guided Residual Encoding}\label{sec:encoding}

\begin{figure}[t]
\centering
\includegraphics[width=\linewidth,trim=21 11.5 0 0,clip]{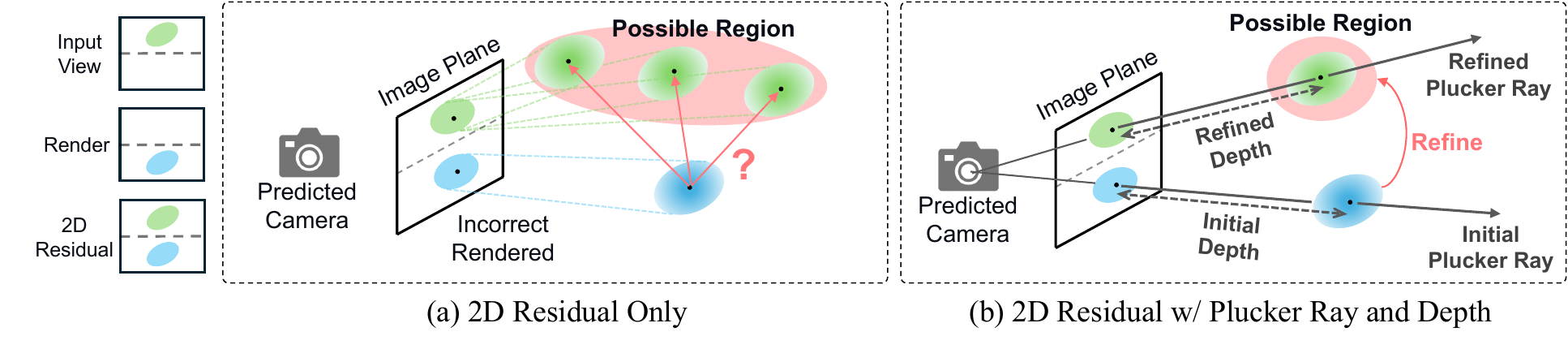}
\caption{\textbf{Geometry-guided residual encoding.} (a)~The 2D residual alone underdetermines the
position update $\Delta\mu$: displacements along the viewing
ray yield identical residuals, leaving a large possible region
(red). (b)~The per-pixel Pl\"ucker ray and rendered depth
encode the line of sight and the current prediction's position
along it, a prior on the current geometry under which the
possible region contracts.}
\label{fig:geom}
\end{figure}

Inferring a 3D positional update from 2D image-space residuals is inherently underdetermined. Because a 2D residual constrains corrections only within the image plane, displacing a Gaussian along its viewing ray yields an identical 2D projection, leaving the residual unchanged regardless of 3D spatial displacement (Figure~\ref{fig:geom} (a)). To resolve this spatial
ambiguity, we explicitly condition the refinement on the 3D
geometry of the current prediction by incorporating per-pixel
Pl\"ucker rays and rendered depth, which describe the line of
sight and the current prediction's position along it
(Figure~\ref{fig:geom} (b)). This geometric context acts as a
soft prior that regulates the direction and scale of the
predicted corrections.

For each context view, we construct a 13-channel feature map by
concatenating the image residual $r_i$, the current rendering
$\hat{I}_i$, the Pl\"ucker-ray map $\ell_i$, and the depth map
$d_i$ rendered alongside $\hat{I}_i$ by the same rasterizer.
The entry $\ell_{i,u}$ of $\ell_i$ at pixel $u = (x, y)$ is
given by
\begin{equation}
\ell_{i,u} = \big(\mathbf{d}_{i,u},\;
\mathbf{o}_i \times \mathbf{d}_{i,u}\big) \in \mathbb{R}^6,
\end{equation}
where $\mathbf{o}_i \in \mathbb{R}^3$ is the center of camera
$p_i$ and $\mathbf{d}_{i,u} \in \mathbb{R}^3$ the unit
direction of the ray from $\mathbf{o}_i$ through pixel $u$.

Each input carries distinct information: the residual $r_i$
measures the per-pixel photometric discrepancy, $\ell_i$
specifies for each pixel the line of sight along which the
positional correction is ambiguous, and $d_i$ locates the
current prediction along this line. Conditioning on $\ell_i$
and $d_i$ thus exposes the geometric structure of the ambiguity
directly in the input, rather than requiring the network to
recover it from appearance. Furthermore, the rays and the depth are computed from exactly the cameras used to place and render the Gaussians. Whether these cameras are externally calibrated or estimated by the backbone, all inputs to $h_\phi$ are thus
expressed in the coordinate frame of the current prediction:
an error in the estimated cameras displaces the residual and
the geometric channels together, leaving them mutually
consistent. Refinement therefore remains well-defined in the
pose-free setting, where no ground-truth cameras are available.

\subsection{Delta Decoder}\label{sec:decoder}

The decoder transforms the geometry-guided features into
per-Gaussian updates in two stages: an image-space dual-branch
mixer followed by per-attribute heads.

\paragraph{Dual-branch mixer.}
The update requires spatial context at two scales: per-pixel
detail at full resolution and broader structure that only a
wide receptive field provides. The mixer must also remain
fast, because it runs at every refinement iteration and its
cost enters the total inference time $K$ times. It therefore
processes the encoding through two branches operating at
different resolutions. The coarse branch rearranges the input
from $H \times W \times C$ to
$\tfrac{H}{2} \times \tfrac{W}{2} \times 4C$ via
pixel-unshuffle~\citep{7780576}, a lossless space-to-depth reshaping, and
applies dilated convolutions~\citep{yu2016multiscale} at this half resolution; the
number of spatial positions drops to a quarter while dilation
further enlarges the receptive field, so the branch covers
broad structural context at a fraction of the full-resolution
cost. The detail branch instead stays at the full
$H \times W$ resolution and applies a few lightweight
convolutions, preserving the high-frequency residual signal.
The two branch outputs are fused additively into a single
feature map, from which the per-Gaussian features of the next
stage are sampled. The resulting mixer surpasses a full-resolution single-branch
variant in quality at $1.84\times$ lower time overhead, since
most of its computation runs on the quarter-area grid.

\paragraph{Per-attribute head.}
To map the image-space features to Gaussian updates, each
Gaussian samples the fused feature map at its projected mean in
each context view, and the per-view samples are aggregated by
an average weighted by the Gaussian's rasterizer blending
weights. The aggregated feature is concatenated with the
Gaussian's current parameters and decoded into the update. The
position, opacity, and color updates differ substantially in
numerical scale, so decoding them with a single shared MLP lets
the largest signal dominate the others; we therefore employ
three parallel MLPs, one per attribute. All output layers are
zero-initialized so that refinement starts exactly at
$\Delta = 0$~\citep{zhang2023controlnet}.

\begin{table}[t]
\caption{Main results on the DL3DV testset (140 scenes,
$224^2$) at 6 / 12 / 24 context views. $p$, $K$: \checkmark{}
means the ground-truth pose / intrinsics are supplied, blank means
predicted by the model.}
\label{tab:main}
\centering
\small
\setlength{\tabcolsep}{3pt}
\resizebox{\textwidth}{!}{%
\begin{tabular}{lccccccccccc}
\toprule
& & & \multicolumn{3}{c}{6 views} & \multicolumn{3}{c}{12 views} & \multicolumn{3}{c}{24 views} \\
\cmidrule(lr){4-6}\cmidrule(lr){7-9}\cmidrule(lr){10-12}
Method & $p$ & $K$ & PSNR$\uparrow$ & SSIM$\uparrow$ & LPIPS$\downarrow$ & PSNR$\uparrow$ & SSIM$\uparrow$ & LPIPS$\downarrow$ & PSNR$\uparrow$ & SSIM$\uparrow$ & LPIPS$\downarrow$ \\
\midrule
MVSplat~\citep{chen2024mvsplat} & \checkmark & \checkmark & 22.66 & 0.760 & 0.173 & 21.29 & 0.709 & 0.224 & 19.98 & 0.662 & 0.269 \\
DepthSplat~\citep{xu2025depthsplat} & \checkmark & \checkmark & 23.42 & 0.797 & 0.136 & 21.91 & 0.753 & 0.179 & 20.09 & 0.690 & 0.240 \\
YoNoSplat~\citep{ye2026yonosplat} & \checkmark & \checkmark & 24.72 & 0.817 & 0.139 & 23.29 & 0.773 & 0.177 & 22.67 & 0.758 & 0.192 \\
\textbf{DeltaSplat} & \checkmark & \checkmark & \textbf{26.70} & \textbf{0.866} & \textbf{0.113} & \textbf{25.33} & \textbf{0.833} & \textbf{0.142} & \textbf{24.72} & \textbf{0.824} & \textbf{0.151} \\
\midrule
NoPoSplat~\citep{ye2025noposplat} & & \checkmark & 22.77 & 0.743 & 0.179 & 19.38 & 0.563 & 0.318 & 17.86 & 0.495 & 0.397 \\
YoNoSplat~\citep{ye2026yonosplat} & & \checkmark & 24.89 & 0.819 & 0.138 & 23.15 & 0.758 & 0.183 & 22.35 & 0.731 & 0.205 \\
\textbf{DeltaSplat} & & \checkmark & \textbf{26.64} & \textbf{0.862} & \textbf{0.115} & \textbf{25.05} & \textbf{0.820} & \textbf{0.148} & \textbf{24.21} & \textbf{0.795} & \textbf{0.167} \\
\midrule
AnySplat~\citep{jiang2025anysplat} & & & 19.03 & 0.554 & 0.235 & 18.94 & 0.549 & 0.262 & 19.70 & 0.596 & 0.249 \\
YoNoSplat~\citep{ye2026yonosplat} & & & 24.53 & 0.804 & 0.142 & 22.93 & 0.746 & 0.187 & 22.17 & 0.720 & 0.209 \\
\textbf{DeltaSplat} & & & \textbf{25.69} & \textbf{0.831} & \textbf{0.126} & \textbf{24.34} & \textbf{0.791} & \textbf{0.159} & \textbf{23.66} & \textbf{0.771} & \textbf{0.177} \\
\bottomrule
\end{tabular}}
\end{table}

\subsection{Training and Inference}\label{sec:impl}\label{sec:training}

The backbone $f_\theta$ and the refinement module $h_\phi$ are
trained jointly in an end-to-end manner by supervising the
output of the final refinement iteration: the loss is computed
after all $K$ refinement iterations, and its gradients
propagate through the rasterizer and the entire unrolled
sequence of updates. The objective combines a pixel-wise mean
squared error with a perceptual
LPIPS~\mbox{\citep{zhang2018lpips}} term over all rendered views:
\begin{equation}
\label{eq:loss}
\mathcal{L}
= \sum_{i \in \mathcal{C} \cup \mathcal{T}}
\Big( \big\| \hat{I}_i^{(K)} - I_i \big\|_2^2
+ \lambda_\text{lp}\,
\text{LPIPS}\big(\hat{I}_i^{(K)},\, I_i\big) \Big),
\end{equation}
where $\hat{I}_i^{(K)}$ is the rendering of the final Gaussian
set $\mathcal{G}^{(K)}$ at camera $p_i$ and $\|\cdot\|_2^2$
denotes the mean squared error over pixels. The context terms
anchor the refined Gaussians to the observed input images,
whereas the target terms supervise generalization to novel
viewpoints. We set $\lambda_\text{lp} = 0.1$ with the VGG
backbone.

Training runs for 50\,k steps with $K{=}3$ and shared weights
across iterations. At test time, $h_\phi$ is applied $K{=}3$
times to its own output, consistent with training.

\section{Experiments}\label{sec:experiments}

\subsection{Experimental Setup}\label{sec:setup}

\paragraph{Dataset.}
We train on the DL3DV-10K~\citep{ling2024dl3dv} trainset
and evaluate on 140-scene testset. All images are
processed at a resolution of $224 \times 224$, and experiments
are conducted with 6, 12, and 24 context views.

\paragraph{Model configuration.}
We use YoNoSplat~\citep{ye2026yonosplat}, pre-trained from a
$\pi^3$ model~\citep{yang2026pi3}, as the backbone. Our model is
initialized from its official DL3DV-trained checkpoint and
trained for 50\,k steps with a batch size of one scene per step on a single NVIDIA RTX PRO 6000 GPU.

\paragraph{Evaluation metrics.}
We report PSNR, SSIM~\citep{wang2004ssim}, and
LPIPS~\citep{zhang2018lpips} (VGG), averaged over the test
scenes. We follow the evaluation protocol of
YoNoSplat~\citep{ye2026yonosplat}, where target camera poses in
the pose-free settings are estimated by photometric optimization
before rendering. Unless otherwise specified, results use the
refined Gaussians after $K{=}3$ iterations.

\subsection{Main Results}\label{sec:main-results}

\paragraph{Pose-free comparison.}
Table~\ref{tab:main} summarizes the main results. In the pose-free
setting with known intrinsics, DeltaSplat reaches 26.64\,dB PSNR
at 6 views, improving its feed-forward backbone by $+1.75$\,dB,
with consistent margins of $+1.90$ and $+1.86$\,dB at 12 and 24
views; SSIM and LPIPS improve accordingly. Notably, DeltaSplat
also surpasses every baseline supplied with ground-truth cameras
(26.64 vs.\ at best 24.72\,dB at 6 views): the gain from
refinement exceeds what the backbone gains from access to exact
cameras.

\paragraph{Fully uncalibrated setting.}
When the intrinsics are also predicted, DeltaSplat attains
25.69\,dB at 6 views, again above every baseline supplied with
ground-truth cameras, and the margin over the backbone widens
with more views ($+1.16$ to $+1.49$\,dB).

\begin{figure}[t]
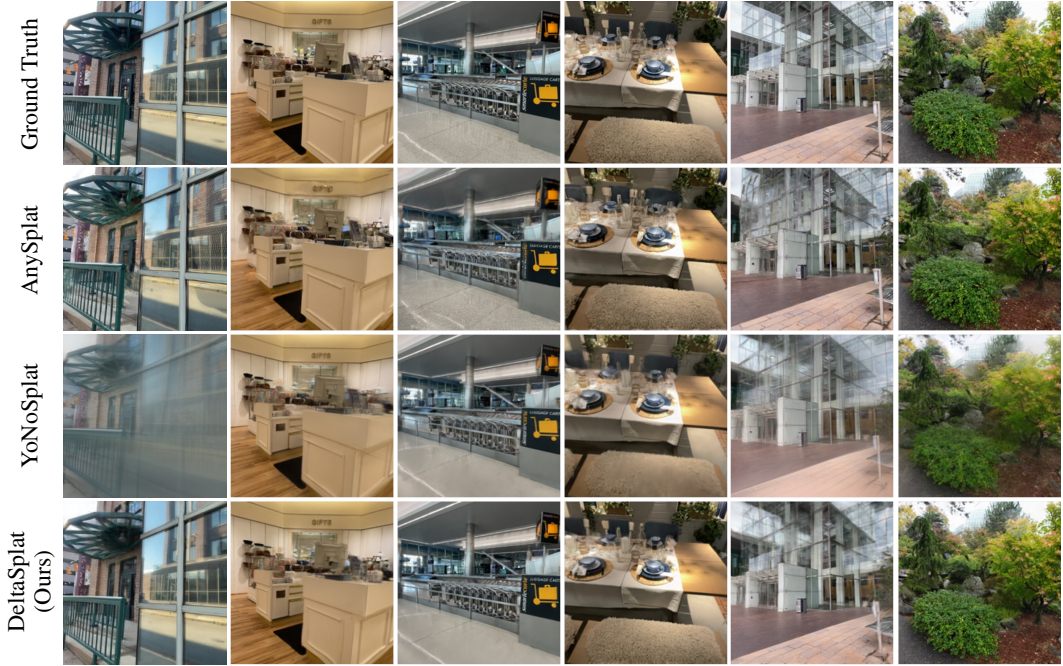

\centering
\setlength{\qcell}{\dimexpr(\linewidth-22.5pt)*450/2750\relax}%
\setlength{\qgap}{\dimexpr\qcell*10/450\relax}%
\setlength{\qcolh}{\dimexpr\qcell*1830/450\relax}%
\savebox{\qualimg}{%
\qscenezero\hspace{\qgap}%
\qgridcol{576}{1387}\hspace{\qgap}%
\qgridcol{1037}{927}\hspace{\qgap}%
\qgridcol{1497}{467}\hspace{\qgap}%
\qgridcol{117}{1847}\hspace{\qgap}%
\qgridcol{1957}{7}%
}%
\begin{minipage}[b]{20pt}\offinterlineskip\centering
\qrowlabel{Ground Truth}\par
\qrowlabel{AnySplat}\par
\qrowlabel{YoNoSplat}\par
\qrowlabel{DeltaSplat\\(Ours)}\par
\kern0pt
\end{minipage}\hspace{2pt}%
\usebox{\qualimg}
\caption{Qualitative results on DL3DV scenes.}
\label{fig:qual}
\end{figure}

\paragraph{Qualitative results.}
Figure~\ref{fig:qual} compares DeltaSplat with the baselines on
DL3DV scenes.

\begin{table}[t]
\caption{Cross-dataset generalization to ScanNet++~\citep{yeshwanth2023scannet++} at $224^2$
with 32 / 64 / 128 context views and known intrinsics.
All models are trained on DL3DV.}
\label{tab:scannet}
\centering
\tablefont
\setlength{\tabcolsep}{4pt}
\begin{tabular*}{\textwidth}{@{\extracolsep{\fill}}lccccccccc}
\toprule
& \multicolumn{3}{c}{32 views} & \multicolumn{3}{c}{64 views} & \multicolumn{3}{c}{128 views} \\
\cmidrule(lr){2-4}\cmidrule(lr){5-7}\cmidrule(lr){8-10}
Method & PSNR$\uparrow$ & SSIM$\uparrow$ & LPIPS$\downarrow$ & PSNR$\uparrow$ & SSIM$\uparrow$ & LPIPS$\downarrow$ & PSNR$\uparrow$ & SSIM$\uparrow$ & LPIPS$\downarrow$ \\
\midrule
YoNoSplat & 17.94 & 0.659 & 0.380 & 18.83 & 0.688 & 0.342 & 19.28 & 0.701 & 0.325 \\
\textbf{DeltaSplat} & \textbf{18.62} & \textbf{0.670} & \textbf{0.371} & \textbf{19.76} & \textbf{0.711} & \textbf{0.328} & \textbf{20.62} & \textbf{0.741} & \textbf{0.298} \\
\bottomrule
\end{tabular*}

\end{table}

\begin{figure}[t]
\captionsetup{justification=raggedright,singlelinecheck=false}%
\setlength{\rtkgap}{2pt}%
\setlength{\rtkcell}{\dimexpr(0.53\linewidth-22pt-3\rtkgap)/4\relax}%
\begin{minipage}[t]{0.53\linewidth}
\vspace{0pt}
\centering
\sbox{\rtkimg}{\vbox{\offinterlineskip\rtkrow{gt}\vskip\rtkgap\rtkrow{yonosplat}\vskip\rtkgap\rtkrow{deltasplat}}}%
\mbox{\vbox{\offinterlineskip\rtklabel{Ground\\Truth}\vskip\rtkgap\rtklabel{YoNoSplat}\vskip\rtkgap\rtklabel{DeltaSplat\\(Ours)}}\hspace{2pt}%
\usebox{\rtkimg}}\par
\captionof{figure}{Qualitative results on RE10K (6 context views, pose-free).}
\label{fig:re10k}
\end{minipage}\hfill
\begin{minipage}[t]{0.45\linewidth}
\vspace{0pt}
\centering
\captionof{table}{Generalization to RE10K~\citep{zhou2018stereo} ($224^2$, 6 context views).
Baselines are trained on RE10K, whereas \mbox{DeltaSplat} is
transferred zero-shot from DL3DV.
$p$, $K$: \checkmark{} means the ground-truth pose / intrinsics
are supplied.}
\label{tab:re10k}
\small
\setlength{\tabcolsep}{3pt}
\begin{tabular*}{\linewidth}{@{\extracolsep{\fill}}lccccc}
\toprule
Method & $p$ & $K$ & PSNR$\uparrow$ & SSIM$\uparrow$ & LPIPS$\downarrow$ \\
\midrule
DepthSplat & \checkmark & \checkmark & 24.16 & 0.846 & 0.145 \\
YoNoSplat & \checkmark & \checkmark & 25.04 & 0.848 & 0.134 \\
\midrule
NoPoSplat & & \checkmark & 22.18 & 0.750 & 0.207 \\
YoNoSplat & & \checkmark & 25.40 & 0.857 & 0.131 \\
\textbf{DeltaSplat} & & \checkmark & \textbf{26.30} & \textbf{0.865} & \textbf{0.129} \\
\midrule
YoNoSplat & & & 24.57 & 0.823 & 0.144 \\
\textbf{DeltaSplat} & & & \textbf{24.87} & \textbf{0.830} & \textbf{0.142} \\
\bottomrule
\end{tabular*}
\end{minipage}
\end{figure}

\paragraph{Efficiency.}
The refinement module adds ${\sim}2.2\%$ to the backbone
parameters, and the $K{=}3$ refinement iterations add
${\sim}0.11$\,s per scene on top of the backbone forward pass.
By contrast, the optional post-optimization of
YoNoSplat~\citep{ye2026yonosplat} requires 17.7--165\,s per scene
depending on the view count, while DeltaSplat remains entirely
feed-forward.

\subsection{Ablation Study}\label{sec:ablations}

\begin{figure}[t]
\centering
\includegraphics[width=\linewidth,trim=11.75 6 12 11.75,clip]{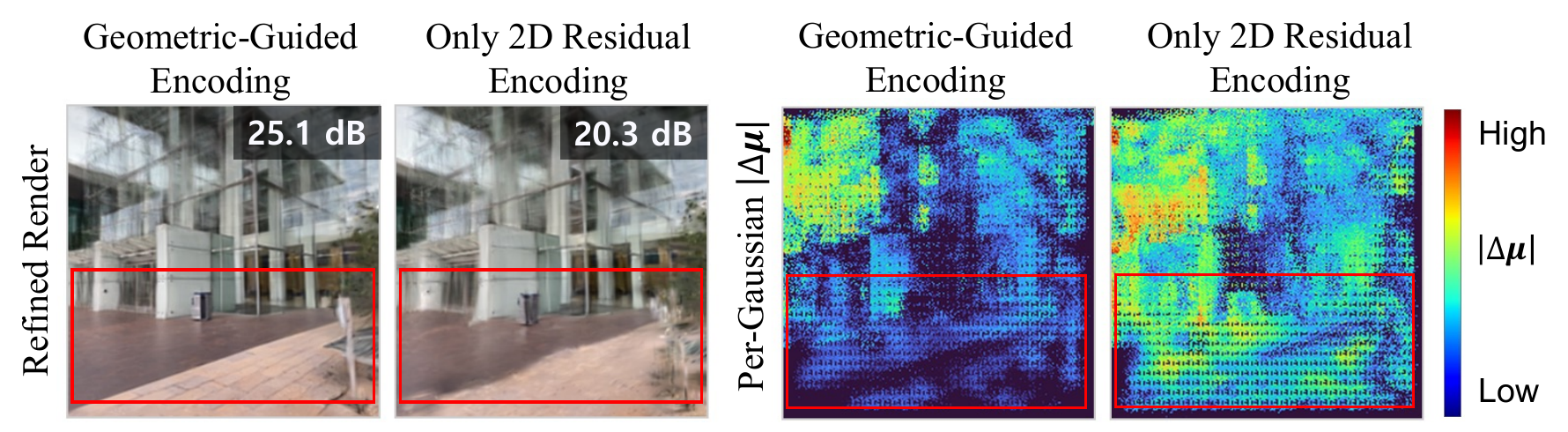}
\caption{\textbf{Effect of the 3D geometry prior.}
Each pair compares the geometry-aimed encoding with an encoding
built from the 2D residual alone; the left pair shows the refined
render and the right pair the per-Gaussian $|\Delta\mu|$.
Without the Pl\"ucker-ray and depth channels the same residual
drives larger and more scattered position updates, and PSNR
falls from $25.1$ to $20.3$\,dB.}
\label{fig:hevidence}
\end{figure}

\begin{table}[t]
\caption{Ablations on the DL3DV test set in the pose-free setting with known
intrinsics (6 context views). (a)~Each row removes a single component from the
full model. (b)~Effect of the number of refinement iterations $K$ applied at test
time, with per-scene refinement time; $K{=}0$ denotes the unrefined backbone
output, and the model is trained with $K{=}3$.}
\label{tab:ablation}
\sbox{\tabA}{\small\renewcommand{\arraystretch}{1.15}\setlength{\tabcolsep}{1.5pt}%
\begin{tabular}{lccc}
\toprule
Variant & PSNR$\uparrow$ & SSIM$\uparrow$ & LPIPS$\downarrow$ \\
\midrule
Full model & \textbf{26.64} & \textbf{0.862} & \textbf{0.115} \\
\makecell[l]{w/o geometric\\ conditioning} & 25.27 & 0.823 & 0.146 \\
\makecell[l]{w/o dual-branch mixer\\ \scriptsize(single-branch mixer)} & 26.05 & 0.846 & 0.132 \\
\makecell[l]{w/o per-attribute MLP\\ \scriptsize(single MLP)} & 26.22 & 0.851 & 0.127 \\
\bottomrule
\end{tabular}}%
\def\tabBbody{\begin{tabular}{lccccccc}
\toprule
 & Backbone & \multicolumn{6}{c}{Refinement iterations} \\
\cmidrule(lr){2-2}\cmidrule(lr){3-8}
$K$ & 0 & 1 & 2 & 3 & 4 & 5 & 6 \\
\midrule
PSNR$\uparrow$ & 24.80 & 26.02 & 26.45 & \textbf{26.64} & 26.48 & 26.32 & 26.17 \\
SSIM$\uparrow$ & 0.818 & 0.847 & 0.857 & \textbf{0.862} & 0.858 & 0.853 & 0.848 \\
LPIPS$\downarrow$ & 0.140 & 0.126 & 0.118 & \textbf{0.115} & 0.119 & 0.124 & 0.130 \\
Time (ms)$\downarrow$ & 54 & 90 & 125 & \textbf{160} & 194 & 230 & 266 \\
\bottomrule
\end{tabular}}%
\sbox{\tabB}{\small\renewcommand{\arraystretch}{1.15}\setlength{\tabcolsep}{3pt}\tabBbody}%
\setlength{\tabdh}{\dimexpr(\ht\tabA+\dp\tabA-\ht\tabB-\dp\tabB)/6\relax}%
\sbox{\tabB}{\small\setstretchpt{1.15pt+\tabdh/10}\setlength{\tabcolsep}{3pt}\tabBbody}%
\setlength{\tabAw}{\dimexpr(\linewidth-8pt)*\wd\tabA/(\wd\tabA+\wd\tabB)\relax}%
\setlength{\tabBw}{\dimexpr\linewidth-8pt-\tabAw\relax}%
\begin{subtable}[t]{\tabAw}
\centering
\caption{Component ablations}
\label{tab:component}
\resizebox{\linewidth}{!}{\usebox{\tabA}}
\end{subtable}\hfill
\begin{subtable}[t]{\tabBw}
\centering
\caption{Number of refinement iterations $K$}
\label{tab:kiter}
\resizebox{\linewidth}{!}{\usebox{\tabB}}
\end{subtable}
\end{table}

\paragraph{Geometric conditioning.}
Omitting the Pl\"ucker-ray and depth channels forces the network to infer 3D corrections solely from 2D features, with two consequences shown in Table~\ref{tab:component}. First, overall visual quality degrades: removing the geometric channels reduces PSNR by $1.37$\,dB, canceling about three quarters of the $1.84$\,dB gain that refinement provides over the unrefined backbone output.
Second, positional updates lose spatial directionality: the magnitude of spatial corrections $|\Delta\mu|$ becomes substantially larger and more dispersed (Figure~\ref{fig:hevidence}), indicating that corrections applied without explicit geometric frames lack directional precision. The geometric channels thus supply exactly the information that the 2D residual underdetermines: the line of sight of each pixel and the current prediction's position along it.

\paragraph{Mixer and head design.}
Table~\ref{tab:component} also compares the dual-branch mixer against a single-branch variant that processes the encoding entirely at full resolution, and the per-attribute heads against a single shared MLP. Removing the dual-branch design costs $0.59$\,dB, as broad structural context must then be gathered entirely at full resolution. Collapsing the per-attribute heads into a single MLP reduces PSNR by $0.42$\,dB, since the update channels differ in numerical scale and a shared decoder lets the largest signal dominate.

\paragraph{Refinement iterations.}
Table~\ref{tab:kiter} varies the number of refinement iterations
applied at test time. Quality improves monotonically over the
first three iterations and peaks at the training setting
$K{=}3$, while the refinement time grows linearly from 54\,ms
for the unrefined backbone output to 266\,ms at $K{=}6$.
Iterating beyond the training horizon degrades quality only
gradually without destabilizing the prediction, and yields no
further gains, so $K{=}3$ offers the best quality--cost
trade-off.

\paragraph{Cross-dataset generalization.}
Tables~\ref{tab:scannet} and~\ref{tab:re10k} evaluate DeltaSplat
beyond the DL3DV training distribution. On
ScanNet++~\citep{yeshwanth2023scannet++}, the DL3DV-trained
models are evaluated without any fine-tuning, using 32, 64, and
128 context views with known intrinsics. DeltaSplat outperforms
the backbone at every view count, and the margin grows from
$+0.68$\,dB at 32 views to $+1.34$\,dB at 128 views, following
the same trend as on DL3DV. On RE10K~\citep{zhou2018stereo} with
6 context views, the baselines are trained on RE10K while
DeltaSplat is transferred zero-shot from DL3DV. DeltaSplat
attains 26.30\,dB with known intrinsics and 24.87\,dB in the
fully uncalibrated setting, surpassing the backbone in both
settings (Figure~\ref{fig:re10k}). The refinement thus transfers to unseen datasets
without any adaptation.

\section{Conclusion}\label{sec:conclusion}

We introduced DeltaSplat, a lightweight Gaussian refinement
module for pose-free feed-forward 3DGS. Conditioning image-space
residuals on per-pixel Pl\"ucker rays and rendered depth, it
lifts 2D evidence into 3D updates that progressively refine the
position, opacity, and color of the predicted Gaussians. With
${\sim}2.2\%$ additional parameters and a fully feed-forward
inference path, DeltaSplat consistently improves a
state-of-the-art backbone across camera regimes, view counts,
and datasets, offering a general plug-in quality stage for
pose-free feed-forward reconstruction. A natural next step is to
extend the refinement to higher resolutions and to other
feed-forward backbones.

\newpage %
\bibliography{references}
\bibliographystyle{iclr2027_conference}

\end{document}

%% file: math_commands.tex
\usepackage{amsmath,amsfonts,bm}

\def\eqref#1{equation~\ref{#1}}

\def\1{\bm{1}}

\DeclareMathAlphabet{\mathsfit}{\encodingdefault}{\sfdefault}{m}{sl}
\SetMathAlphabet{\mathsfit}{bold}{\encodingdefault}{\sfdefault}{bx}{n}

